\pdfoutput=1

\documentclass[11pt]{article}

\usepackage[]{acl}

\usepackage{times}
\usepackage{latexsym}
\usepackage{lipsum}
\usepackage{booktabs}
\usepackage[labelfont=bf]{caption}
\usepackage{multirow}
\usepackage{graphicx}
\usepackage{amssymb}
\usepackage{amsmath}
\usepackage{soul}

\usepackage[T1]{fontenc}

\usepackage[utf8]{inputenc}

\usepackage{microtype}

\usepackage{inconsolata}

\usepackage{graphicx} 
\usepackage{multirow}
\parskip=\medskipamount
\newcommand*\rot{\rotatebox{90}}
\newcommand{\fl}{{\tt fluency}}
\newcommand{\coh}{{\tt coherence}}
\newcommand{\rel}{{\tt relevance}}
\newcommand{\fa}{{\tt faithfulness}}
\newcommand{\asp}{{\tt aspect coverage}}
\newcommand{\sent}{{\tt sentiment consistency}}
\newcommand{\spec}{{\tt specificity}}

\newcommand{\mistral}[1]{{\tt Mistral-$#1$B}}

\newcommand{\llamatwo}[2]{{\tt LLaMA$#1$-$#2$B}}
\newcommand{\vicuna}[1]{{\tt Vicuna-$#1$B}}
\newcommand{\zephyr}[1]{{\tt Zephyr-$#1$B}}
\newcommand{\solar}[1]{{\tt Solar-$#1$B}}
\newcommand{\gpt}[1]{{\tt GPT-$#1$}}
\newcommand{\chatgpt}[1]{{\tt ChatGPT-$#1$}}

\newcommand{\bmistral}[1]{{\tt \textbf{Mistral-$\mathbf{#1}$B}}}

\newcommand{\bllamatwo}[2]{{\tt \textbf{LLaMA$\mathbf{#1}$-$\mathbf{#2}$B}}}
\newcommand{\bvicuna}[1]{{\tt \textbf{Vicuna-$\mathbf{#1}$B}}}
\newcommand{\bzephyr}[1]{{\tt \textbf{Zephyr-$\mathbf{#1}$B}}}
\newcommand{\bsolar}[1]{{\tt \textbf{Solar-$\mathbf{#1}$B}}}
\newcommand{\bgpt}[1]{{\tt \textbf{GPT-$\mathbf{#1}$}}}
\newcommand{\bchatgpt}[1]{{\tt \textbf{ChatGPT-$\mathbf{#1}$}}}

\newcommand\blfootnote[1]{%
  \begingroup
  \renewcommand\thefootnote{}\footnote{#1}%
  \addtocounter{footnote}{-1}%
  \endgroup
}

\newcommand{\emojilorring}{\raisebox{-0.2em}{\includegraphics[height=1.6em]{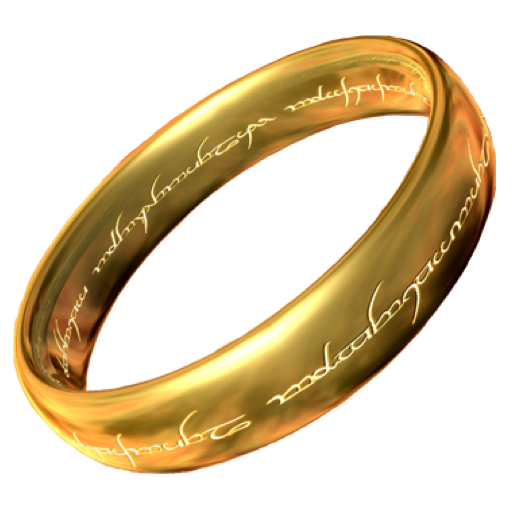}}}

\title{\emojilorring{}\hspace{-0.2em}ne Prompt To Rule Them All: LLMs for Opinion Summary Evaluation}

\author{Tejpalsingh Siledar$^\dag$$^\clubsuit$,
Swaroop Nath$^\dag$$^\clubsuit$, Sri Raghava$^\dag$$^\clubsuit$, Rupasai Rangaraju$^\dag$$^\clubsuit$, 
\\ 
\textbf{Swaprava Nath$^\clubsuit$, Pushpak Bhattacharyya$^\clubsuit$, Suman Banerjee$^\spadesuit$, Amey Patil$^\spadesuit$,}\\
\textbf{Sudhanshu Shekhar Singh$^\spadesuit$, Muthusamy Chelliah$^\spadesuit$, Nikesh Garera$^\spadesuit$}\\
        $^\clubsuit$Computer Science and Engineering, IIT Bombay, India, 
        $^\spadesuit$Flipkart, India \\
        \texttt{\{tejpalsingh, swaroopnath, sriraghava, rupasai, swaprava,
        pb\}@cse.iitb.ac.in}
        }

\begin{document}
\maketitle
\blfootnote{$^\dag$ Equal contribution}
\begin{abstract}
Evaluation of opinion summaries using conventional reference-based metrics often fails to provide a comprehensive assessment and exhibits limited correlation with human judgments. While Large Language Models (LLMs) have shown promise as reference-free metrics for NLG evaluation, their potential remains unexplored for opinion summary evaluation. Furthermore, the absence of sufficient opinion summary evaluation datasets hinders progress in this area. In response, we introduce the \textsc{SummEval-Op} dataset, encompassing $7$ dimensions crucial to the evaluation of opinion summaries: \texttt{fluency}, \texttt{coherence}, \texttt{relevance}, \texttt{faithfulness}, \texttt{aspect coverage}, \texttt{sentiment consistency}, and \texttt{specificity}. We propose \textsc{Op-I-Prompt}, a dimension-independent prompt, along with \textsc{Op-Prompts}, a dimension-dependent set of prompts for opinion summary evaluation. Our experiments demonstrate that \textsc{Op-I-Prompt} emerges as a good alternative for evaluating opinion summaries, achieving an average Spearman correlation of $\mathbf{0.70}$ with human judgments, surpassing prior methodologies. Remarkably, we are the first to explore the efficacy of LLMs as evaluators, both on closed-source and open-source models, in the opinion summary evaluation domain.
\end{abstract}

\section{Introduction}

\begin{figure}[t]
    \centering
    \includegraphics[width=1\columnwidth]{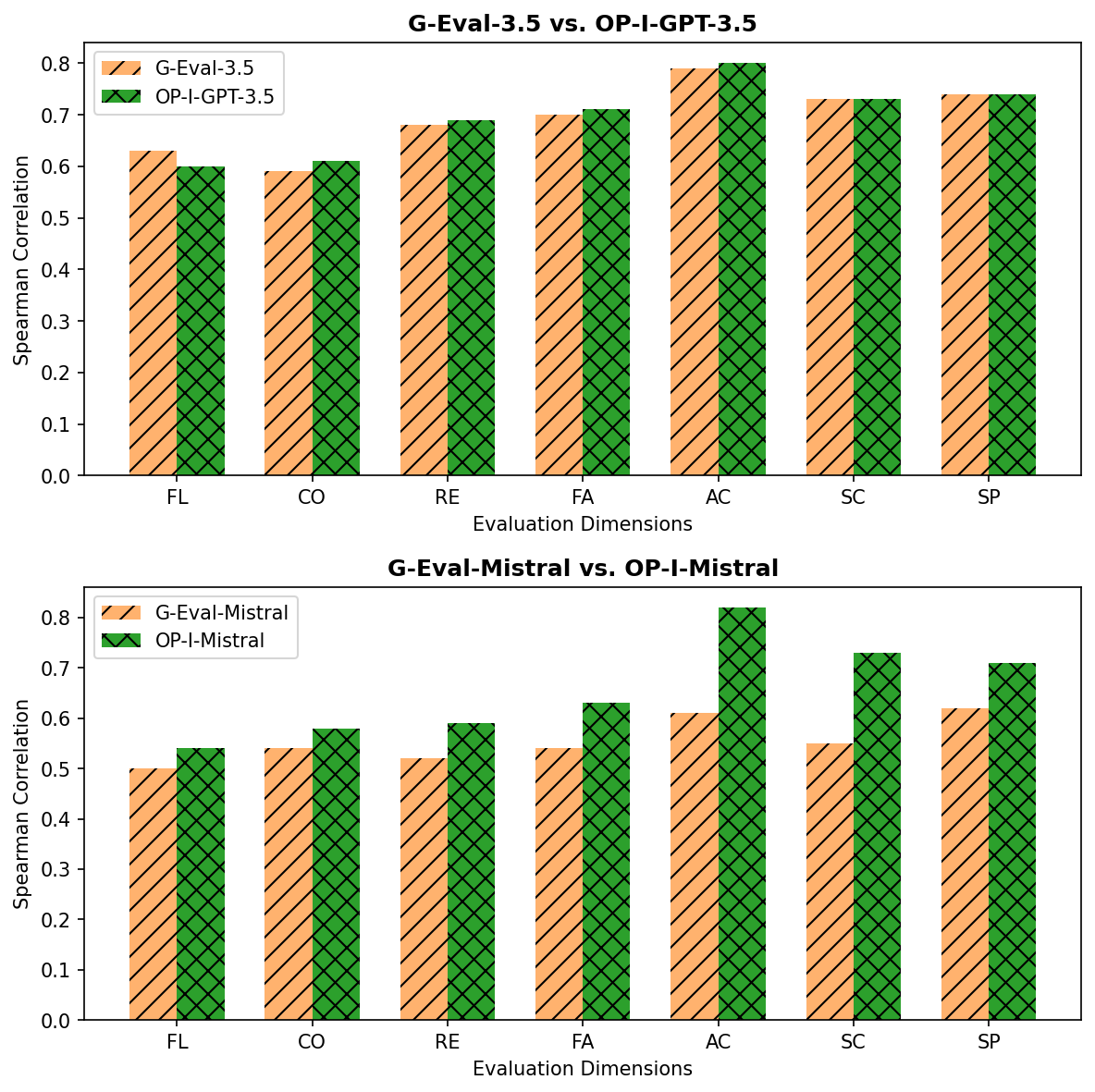}
    \caption{\textbf{\textsc{G-Eval} vs. \textsc{Op-I-Prompt}}. On closed-source model (\chatgpt{3.5}) our \textsc{Op-I-Prompt} shows comparable performance whereas on open-source model (\mistral{7}) our approach outperforms \textsc{G-Eval} on $7$ dimensions: \fl \:(FA), \coh \:(CO), \rel \:(RE), \fa \:(FA), \asp \:(AC), \sent \:(SC), and \spec \:(SP). Check Figure \ref{fig:generation_runs} for more details.} 
    \label{fig:comparison}
\end{figure}
Opinion summarization systems predominantly use traditional metrics such as \textsc{Rouge} \citep{lin-2004-rouge} and \textsc{BertScore} \citep{Zhang2019BERTScoreET} for automatic evaluation, however, they have been shown to have poor correlations with human judgments \citep{shen2023opinsummeval}. Moreover, these metrics fall short of comprehensively evaluating opinion summaries. Additionally, obtaining reference-based datasets at a large scale is an expensive process. 

Recently, Large Language Models (LLMs) have been utilized as reference-free evaluators for Natural Language Generation (NLG) outputs \citep{fu2023gptscore, chiang-lee-2023-large, chiang-lee-2023-closer, wang-etal-2023-chatgpt, liu-etal-2023-g}. The idea is to prompt a powerful LLM such as \chatgpt{3.5}/\gpt{4} to evaluate an output on certain criteria. However, their suitability has not been explored at all for evaluating opinion summaries. Moreover, these approaches have been tested only on closed-source models (\chatgpt{3.5}/\gpt{4}) primarily because of the limitations of the open-source models in following instructions and producing the desired output \citep{chiang-lee-2023-closer}. 

To this end, we first create \textsc{SummEval-Op}, a reference-free opinion summarization dataset covering $7$ dimensions, for the e-commerce domain. Next, we present \textsc{Op-I-Prompt} and \textsc{Op-Prompts} tailored for opinion summary evaluation. We investigate their suitability to both closed-source and open-source models. Experiments reveal that \textsc{Op-I-Prompt} emerges as a good alternative for evaluating opinion summaries across all $7$ dimensions.

Our contributions are:
\begin{enumerate}
    \item \textbf{\textsc{SummEval-Op}}\footnote{\url{https://github.com/tjsiledar/SummEval-OP}}, an opinion summary evaluation benchmark dataset,  consisting of a total of $2,912$ summary annotations, assessing $13$ opinion summaries for $32$ products from the Amazon test set. The evaluation covers $\mathbf{7}$ \textbf{dimensions}- {\tt fluency}, {\tt coherence}, {\tt relevance}, {\tt faithfulness}, {\tt aspect coverage}, {\tt sentiment consistency}, and {\tt specificity}  related to the evaluation of opinion summaries (Section \ref{benchmark_dataset}). 
    \item \textbf{\textsc{Op-I-Prompt}}, a dimension-independent prompt and \textbf{\textsc{Op-Prompts}}, a dimension-dependent set of prompts, enabling opinion summary evaluation for all the $7$ dimensions. Experiments indicate that the \textsc{Op-I-Prompt} generally outperforms existing approaches on both closed-source and open-source models by $\mathbf{9\%}$ on average in correlation with human judgments (Figure \ref{fig:comparison},  Section \ref{method}). To the \textit{best of our knowledge} we are the first to test the applicability of different prompt approaches on open-source LLMs.
    \item Benchmarking of $\mathbf{9}$ recent LLMs (closed and open-source) on the aforementioned $7$ dimensions for the task of opinion summarization, which to the \textit{best of our knowledge} is first of its kind (Table \ref{tab:model_performance}, Section \ref{results_analysis}).
    \item Detailed analysis, comparing an open-source LLM against a closed-source LLM acting as evaluators for automatic evaluation of opinion summaries on $7$ dimensions. Analysis indicates that \textsc{Op-I-Prompt} emerges as a good alternative for evaluating opinion summaries showing a high correlation (spearman correlation of $\mathbf{0.70}$ on average) with humans when compared with alternatives (Section \ref{results_analysis}).
\end{enumerate}

\section{Related Work}

\textbf{LLM-based Evaluators.}\; \citet{fu2023gptscore} introduced GPTScore that operates on the premise that a generative pre-training model (e.g. GPT-3) is likely to assign a higher probability to the generation of high-quality text in line with provided instructions and context. \citet{chiang-lee-2023-large} were the first to explore LLMs for evaluation. \citet{chiang-lee-2023-closer} provide concrete guidelines that improve ChatGPT's correlation with humans. \citet{wang-etal-2023-chatgpt} conducted an initial survey exploring the utilization of ChatGPT as an NLG evaluator. \citet{kocmi-federmann-2023-large} used GPT models for evaluating machine learning tasks. \citet{liu-etal-2023-g} introduced G-Eval, a framework for evaluation of NLG outputs using \textit{Chain of Thought} (CoT) \citep{wei2023chainofthought} and assigning weights to a predetermined set of integer scores based on their generation probabilities from GPT-3/4. \citet{chen2023exploring} were the first to investigate approaches to reference-free NLG evaluation using LLMs, finding that an explicit score generated by ChatGPT is the most effective and stable approach. \citet{zheng2023judging} show that strong LLMs such as GPT-4 achieve a similar level of agreement to that of humans and hence can be used to approximate human preferences. Our work investigates two prompt strategies and tests the applicability of different prompt approaches on closed-source and open-source LLMs for opinion summary evaluation for $7$ dimensions.

\noindent\textbf{Opinion Summary Evaluation Benchmark.} \cite{shen2023opinsummeval} created the \textsc{OpinSummEval} dataset, utilizing the Yelp test set \citep{pmlr-v97-chu19b}, annotating for $4$ dimensions relevant to opinion summary evaluation. Our work enhances this effort by introducing \textsc{SummEval-Op}, which focuses on the e-commerce domain, constructed using the Amazon test set \cite{brazinskas-etal-2020-unsupervised}. Additionally, we collect annotations for $7$ dimensions on the recent LLM summaries, subsequently establishing benchmarks for comparison.

\noindent\textbf{Opinion Summarization}\; Opinion summarization aims to summarize opinions into concise summaries \citep{wang-ling-2016-neural, siledar-etal-2023-aspect}. Recent approaches use unsupervised methods \citep{pmlr-v97-chu19b, brazinskas-etal-2020-unsupervised}, or creating synthetic datasets with pseudo-summaries \citep{brazinskas-etal-2020-unsupervised, amplayo-lapata-2020-unsupervised, amplayo-etal-2021-unsupervised, elsahar-etal-2021-self, im-etal-2021-self, siledar-etal-2023-synthesize, siledar2024product}. Notably, \citet{amplayo-etal-2021-unsupervised} and \citet{im-etal-2021-self} advanced these methods by generating content plans and using multimodal inputs, respectively. \citet{siledar-etal-2023-synthesize} used lexical and semantic similarities to generate synthetic data. In this work, we benchmark the recent LLMs for the task of opinion summarization.  


\begin{figure*}[t]
    \centering
    \includegraphics[width=2\columnwidth]{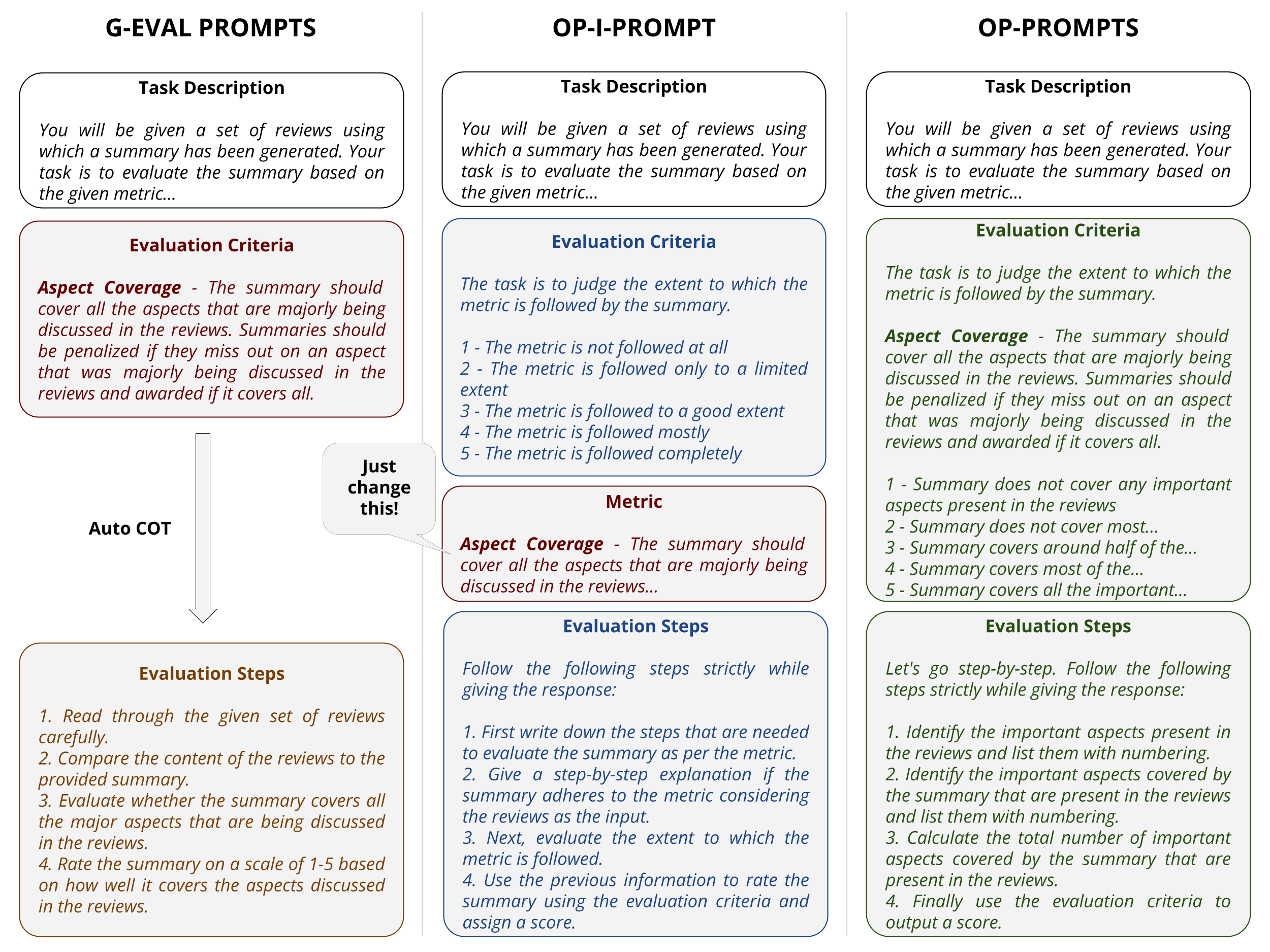}
    \caption{\textbf{Comparison of Prompt Approaches.} \textsc{G-Eval Prompts} first generates the {\tt Evaluation Steps} using {\tt Task Description} and {\tt Evaluation Criteria} in Chain-of-Thought fashion. Finally the full prompt is used to evaluate the opinion summaries. In contrast, our \textbf{\textsc{Op-I-Prompt}} is simpler and has {\tt Task Description}, {\tt Evaluation Criteria}, and {\tt Evaluation Steps} fixed for a dimension/metric independent evaluation. Here, only the {\tt Metric} part needs to be changed for evaluating any dimension/metric. Finally \textbf{\textsc{Op-Prompts}} are dimension/metric dependent prompts that needs to be specifically crafted for each dimension/metric.}
    \label{fig:prompts}
\end{figure*}

\section{Methodology}\label{method}
We describe our dimension independent and dependent prompts and the model scoring function.
\subsection{Prompt Approaches}
Figure \ref{fig:prompts} shows the different prompt approaches for evaluating opinion summaries. In general, the prompts include the following $3$ components-

\noindent(\textit{1}) {\tt \textbf{Task Description}}: Defines the task that the LLM will be performing. In our case, the task is to evaluate a summary corresponding to a set of reviews on a given metric/dimension.

\noindent(\textit{2}) {\tt \textbf{Evaluation Criteria}}: Defines the criteria that will be used to perform the task. In our case, the task being opinion summary evaluation, the criteria is to assign a score ($1-5$) for a certain metric/dimension depending on the extent to which the summary adheres to it.

\noindent(\textit{3}) {\tt \textbf{Evaluation Steps}}: This comprises the steps that the LLM must take to correctly perform the described task. In our case, it contains the steps that the LLM should follow to evaluate a certain metric/dimension.

We propose two prompt approaches: 

\textbf{\textsc{Op-I-Prompt}}\; is a metric-independent opinion summary evaluation prompt. Here we split the {\tt Evaluation Criteria} to create a new component {\tt Metric} consisting only the evaluation dimension. All the remaining components i.e. {\tt Task Description}, {\tt Evaluation Criteria}, and {\tt Evaluation Steps} are crafted in such a way that they are applicable in general to any opinion summary evaluation dimension. This benefits us in the following way: (\textit{a}) we have a metric independent prompt that can now evaluate any metric/dimension just by replacing with the desired definition of the dimension within the {\tt Metric} block (\textit{b}) the remaining components, crafted specifically keeping the task in mind, ensures that the evaluation by LLM takes place as defined by us. 

\textbf{\textsc{Op-Prompts}} is a set of metric-dependent prompts. We specifically handcrafted these prompts for each of the $7$ evaluation dimensions. Although this ensures that the evaluation happens exactly in the way we define, this requires a certain level of expertise in the evaluation domain and prompting. This could be seen as a much stricter version of the prompt compared to \textsc{Op-I-Prompt} where the prompt is suited to any evaluation dimension which is not the case here. A prompt defined for a certain dimension could not be utilized for any other dimension.

In contrast, \textsc{G-Eval} \citep{liu-etal-2023-g} used auto chain-of-thoughts \cite{Wei2022ChainOT} by using {\tt Task Description} and {\tt Evaluation Criteria} to automatically generate the {\tt Evaluation Steps}. Finally, all the components together constitute the \textsc{G-Eval} prompt that is used by an LLM to evaluate summaries. Our work investigates the applicability of all these prompts to both closed-source and open-source models for evaluating opinion summaries.

\subsection{Prompt Design Consideration} 
The design of our prompts was based on the intuition that LLMs would produce improved responses when prompted to justify their evaluations. Our approach ensures that the response reiterates the evaluation metric, highlights both strengths and shortcomings and concludes with an evaluation score based on the criteria outlined in the prompt.

\subsection{Scoring Function}\label{scoring_func}
\citet{liu-etal-2023-g} pointed out the limitation of LLM outputting an integer score and proposed using a weighted average of the scores as the LLMs output, where the weights are the probabilities of the corresponding score. Formally, say, the scoring is scheme is from $\{s_{1},...,s_{j}\}$, the probability of each score $p(s_{k})$ is calculated by an LLM and the final score $o$ is computed as:
\begin{align}
    o = \sum_{k=1}^{j} p(s_{k})\times s_{k}
\end{align}
$p(s_{k})$ for an input $k$ is estimated through an LLM by sampling {\tt n} outputs. In which case, the scoring function just translates to taking a mean over the {\tt n} outputs. We ensure that {\tt n} is large ($\sim100$) to get a reliable estimate of the probabilities. 

\subsection{Evaluation Approach}\label{appendix_correlation}
For each product $d_{i}$ in dataset $\mathcal{D}$, $i \in \{1,...,\mathcal{Z}\}$ we have $\mathcal{N}$ opinion summaries from different models. Let $s_{ij}$ denote the $j^{th}$ summary of the product $d_{i}$, $\mathcal{M}_{m}$ denote the $m^{th}$ evaluation metric, and $\mathcal{K}$ denote the correlation measure. \citet{bhandari-etal-2020-evaluating} defines the summary-level correlation as:
\begin{align}
    \mathcal{R}(a,b) = \frac{1}{\mathcal{Z}} \sum_{i} &\mathcal{K}([\mathcal{M}_{a}(s_{i1}),...,\mathcal{M}_{a}(s_{i\mathcal{N}})], \nonumber \\
    &[\mathcal{M}_{b}(s_{i1}),...,\mathcal{M}_{b}(s_{i\mathcal{N}})])
\end{align}

\section{\textsc{SummEval-Op} Benchmark Dataset} \label{benchmark_dataset}

We created the \textbf{\textsc{SummEval-Op}} benchmark dataset for evaluating the opinion summaries on $7$ dimensions. In this section, we discuss the dataset used, opinion summary evaluation metrics, annotation details, and its analysis.

\subsection{Dataset}\label{amazon_dataset}
We utilized the Amazon test set \citep{He2016UpsAD, brazinskas-etal-2020-unsupervised}, comprising of reviews from $4$ domains: \textit{electronics, home \& kitchen, personal care,} and \textit{clothing, shoes \& jewelry}. The test set contained a total of $32$ products, each with $3$ human-annotated abstractive summaries and $8$ reviews per product. The reviews and summaries are all in English. For our use, we needed one human reference summary per product which we obtained by randomly selecting one of the summaries out of the $3$ for each product. We do not directly consider only one of the human summaries as this would bias the summaries to a single person. 
 

\subsection{Opinion Summarization Metrics}\label{metrics}
The evaluation of opinion summaries focused on the following $7$ dimensions:


\begin{enumerate}
    \item {\tt \textbf{fluency}} \textbf{(FL)}- The quality of summary in terms of grammar, spelling, punctuation, capitalization, word choice, and sentence structure and should contain no errors. The summary should be easy to read, follow, comprehend and should contain no errors. Annotators received specific guidelines on how to penalize summaries based on fluency levels.
    \item {\tt \textbf{coherence}} \textbf{(CO)}- The collective quality of all sentences. The summary should be well-structured and well-organized. The summary should not just be a heap of related information, but should build from sentence to a coherent body of information.
    \item {\tt \textbf{relevance}} \textbf{(RE)}- The summary should not contain opinions that are either not consensus or important. The summary should include only important opinions from the reviews. Annotators were instructed to penalize summaries if they contained redundancies and excess/unimportant information.
    \item {\tt \textbf{faithfulness}} \textbf{(FA)}- Every piece of information mentioned in the summary should be verifiable/supported/inferred from the reviews only. Summaries should be penalized if any piece of information is not verifiable/supported/inferred from the reviews or if the summary overgeneralizes something.
    \item {\tt \textbf{aspect coverage}} \textbf{(AC)}- The summary should cover all the aspects that are majorly being discussed in the reviews. Summaries should be penalized if they miss out on an aspect that was majorly being discussed in the reviews and awarded if it covers all.
    \item {\tt \textbf{sentiment consistency}} \textbf{(SC)}- All the aspects being discussed in the summary should accurately reflect the consensus sentiment of the corresponding aspects from the reviews. Summaries should be penalized if they do not cover accurately the sentiment regarding any aspect within the summary.
    \item {\tt \textbf{specificity}} \textbf{(SP)}- The summary should avoid containing generic opinions. All the opinions within the summary should contain detailed and specific information about the consensus opinions. Summaries should be penalized for missing out details and should be awarded if they are specific.
\end{enumerate}




\subsection{Annotation Details}

For creating the \textbf{\textsc{SummEval-Op}} dataset, annotations were collected for a total of $13$ abstractive summaries per product across $7$ dimensions for $32$ products from the Amazon test set. The $13$ summaries comprised of $1$ human-annotated reference summary (mentioned in Section \ref{amazon_dataset}) and $12$ different model-generated summaries (listed in Section \ref{summarization_models}). To ensure high quality of annotations, each summary was annotated by $3$ raters for $7$ dimensions, amounting to $2,912$ summary-level ratings. Raters were asked to rate the summaries on a scale from $1$ to $5$ (higher is better) along the $7$ dimensions. Each summary for each dimension was rated by $3$ raters. The overall quantity of annotations is:  $3$ (\textit{\# of raters}) x $32$ (\textit{\# of instances}) x $13$ (\textit{\# of summaries}) x $7$ (\textit{\# of dimensions}) = $8,736$ ratings.

We hired $3$ Masters' students with experience in opinion summarization as opposed to crowd workers for the following reasons: \textit{(a)} \citet{gillick-liu-2010-non} demonstrated that summary evaluations from non-experts can significantly diverge from expert annotations and may display inferior inter-annotator agreement, and  \textit{(b)} \citet{fabbri2021summeval} emphasized the significance of employing expert annotators to mitigate quality concerns in human ratings. Similar to \citet{fabbri2021summeval}, we conducted the process in two rounds, to ensure high-quality ratings. In Round II, ratings where the scores of any rater differed from any other rater by $2$ or more points were re-evaluated. The re-evaluation was done through a discussion between the annotators such that ratings from all $3$ differ by at most $1$. We asked the raters to be critical and discuss the ratings during re-evaluation. 

We hired raters who have written papers on opinion summarization (1) or are working in the opinion summarization domain (2). These were male Masters' students aged 21-30. They were provided with detailed guidelines for evaluating summaries on the $7$ dimensions. All $3$ raters received stipends suitable for the tasks. Models associated with summaries were not revealed to the raters to remove any bias.
Check \textbf{Appendix} \ref{rater_agreement}.

\subsection{Annotation Analysis}

\begin{table}[t]
    \centering
    \resizebox{1\columnwidth}{!}{%
    \begin{tabular}{lcc}
         \toprule
        & \textbf{Round-I} $\uparrow$ & \textbf{Round-II} $\uparrow$\\
        \midrule
        \fl & $0.55$ & $0.84$ \\
        \coh & $0.43$ & $0.73$\\
        \rel & $0.50$ & $0.79$ \\
        \fa & $0.63$ & $0.86$ \\
        \asp & $0.64$ & $0.82$ \\
        \sent & $0.41$ & $0.78$\\
        \spec & $0.34$ & $0.76$\\
        \midrule
        \textbf{\textsc{AVG}} & $0.50$ & $0.80$ \\
        \bottomrule
    \end{tabular}
    }
    \caption{\textbf{Krippendorff's alpha coefficient} (${\alpha}$) for Round-I and Round-II on $7$ dimensions. As expected, we see an improvement in Round-II coefficient scores. 
    }
    \label{tab:krip_alpha}
\end{table}

\begin{figure}[t]
    \centering
    \includegraphics[width=1\columnwidth]{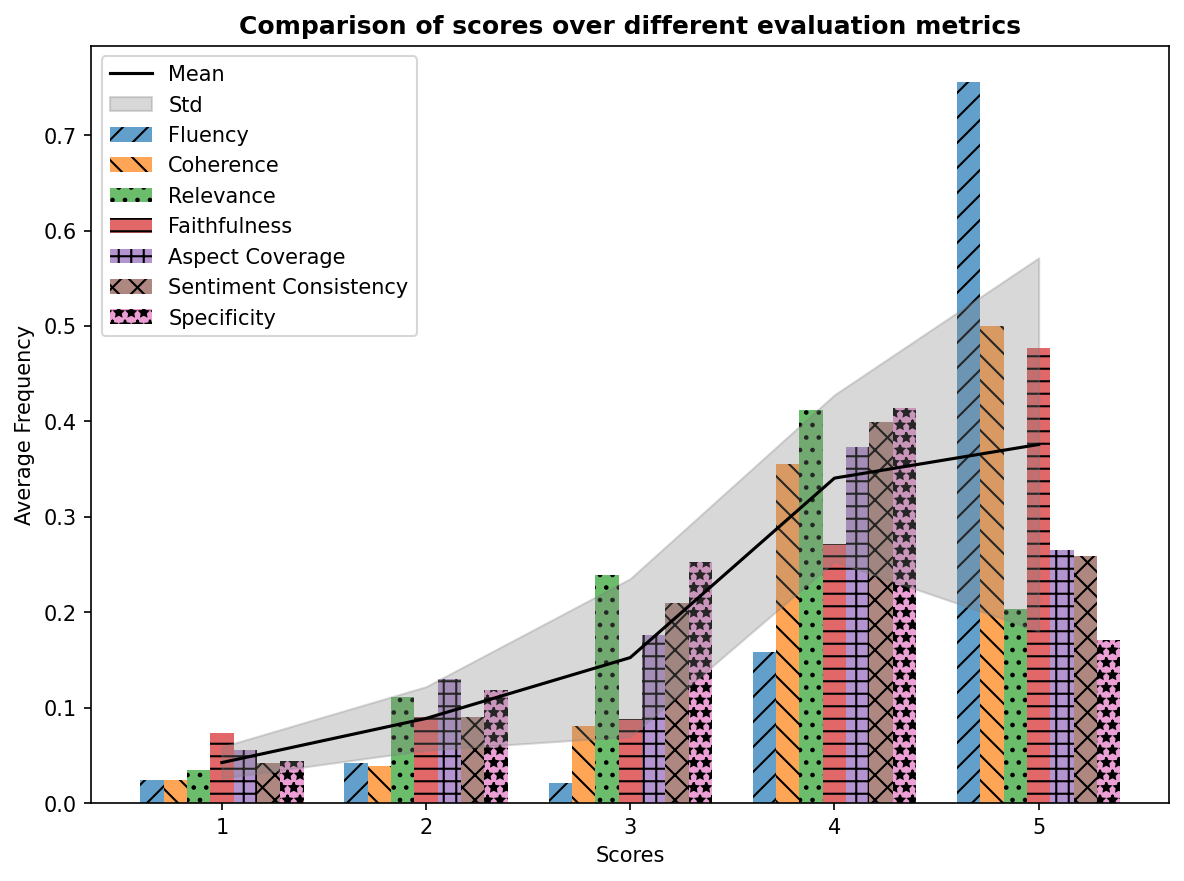}
    \caption{\textbf{Ratings Distribution.} We plot the average frequency of scores obtained by human raters across $7$ dimensions. A score of $4$ or $5$ is mostly preferred.} 
    \label{fig:ratings_plot}
\end{figure}

We evaluated the inter-rater agreement for the $3$ raters using Krippendorff's alpha coefficient ($\alpha$) \citep{Krippendorff2011ComputingKA}. For Round-I, we found the coefficient to be $0.50$ indicating \textit{moderate aggrement} ($0.41 \le \alpha \le 0.60$). For Round-II, the coefficient increased to $0.80$, indicating \textit{substantial agreement} ($0.61 \le \alpha \le 0.80$). We report the dimension-wise agreement scores for both rounds in Table \ref{tab:krip_alpha}. We observe that for both Round-I and Round-II, \fa \:and \asp \:score higher than others. This is mostly because \fa \:and \asp \:could be identified by cross-examining with the reviews. After Round-II, \coh \:and \spec \:are the most disagreed upon between raters. 

Figure \ref{fig:ratings_plot} shows the average frequency of assigning a particular score by human raters for $7$ dimensions. We make some key observations: (\textit{a}) a score of $4$ or $5$ is mostly preferred. This could be attributed to the fact that most of the models are LLMs which are doing pretty well for summary generation tasks. (\textit{b}) for \fl, \coh, and \fa \:a score of $5$ dominates. This indicates that the LLMs are doing good in terms of these dimensions. (\textit{c}) for \rel, \asp, \sent, and \spec \:raters majorly prefer a score of $4$. 


\begin{table*}[t]
    \centering
    \resizebox{2\columnwidth}{!}{%
    \begin{tabular}{clcccccccccccccc}
    \toprule
          & & \multicolumn{2}{c}{\textbf{FL} $\uparrow$}  & \multicolumn{2}{c}{\textbf{CO} $\uparrow$} & \multicolumn{2}{c}{\textbf{RE} $\uparrow$} & \multicolumn{2}{c}{\textbf{FA} $\uparrow$} & \multicolumn{2}{c}{\textbf{AC} $\uparrow$} & \multicolumn{2}{c}{\textbf{SC} $\uparrow$} & \multicolumn{2}{c}{\textbf{SP} $\uparrow$}\\
         \cmidrule(lr){3-4} \cmidrule(lr){5-6} \cmidrule(lr){7-8} \cmidrule(lr){9-10} \cmidrule(lr){11-12} \cmidrule(lr){13-14} \cmidrule(lr){15-16}
         & & $\rho$ & $\tau$ & $\rho$ & $\tau$ & $\rho$ & $\tau$ & $\rho$ & $\tau$ & $\rho$ & $\tau$ & $\rho$ & $\tau$ & $\rho$ & $\tau$ \\  

    \midrule
    \multirow{15}{*}
        {\rot{\textbf{\textsc{SummEval-Op} (Ours) }}} & \textsc{Humans} & $0.80$ & $0.77$ & $0.81$ & $0.76$ & $0.91$ & $0.86$ & $0.89$ & $0.85$ & $0.93$ & $0.87$ & $0.91$ & $0.85$ & $0.92$ & $0.87$ \\
    \cmidrule{2-16}
       &  \textsc{Rouge-1} & $-0.36$ & $-0.28$ & $-0.30$ & $-0.24$ & $-0.31$ & $-0.23$ & $-0.35$ & $-0.26$ & $-0.44$ & $-0.32$ & $-0.38$ & $-0.29$ & $-0.30$ & $-0.23$ \\
        & \textsc{Rouge-2} & $-0.23$ & $-0.18$ & $-0.14$ & $-0.10$ & $-0.17$ & $-0.12$ & $-0.21$ & $-0.16$ & $-0.26$ & $-0.19$ & $-0.24$ & $-0.18$ & $-0.14$ & $-0.09$\\
       & \textsc{Rouge-L} & $-0.39$ & $-0.32$ & $-0.30$ & $-0.23$ & $-0.34$ & $-0.25$ & $-0.40$ & $-0.30$ & $-0.51$ & $-0.37$ & $-0.45$ & $-0.33$ & $-0.38$ & $-0.27$ \\
         & \textsc{BERTScore} & $-0.32$ & $-0.27$ & $-0.28$  & $-0.22$ & $-0.29$ & $-0.22$ & $-0.34$ & $-0.26$ & $-0.51$ & $-0.43$ & $-0.41$ & $-0.33$ & $-0.37$ & $-0.28$ \\
        \cmidrule{2-16}
         & \textsc{BARTScore} & $-0.19$ & $-0.15$ & $-0.19$ & $-0.14$ & $-0.29$ & $-0.22$ & $-0.33$ & $-0.25$ & $-0.45$ & $-0.35$ & $-0.37$ & $-0.28$ & $-0.36$ & $-0.27$ \\
         & \textsc{SummaC} & $0.23$  & $0.20$ & $0.18$ & $0.14$ & $0.30$ & $0.25$ & $0.25$ & $0.21$ & $0.24$ & $0.19$ & $0.25$ & $0.20$ & $0.26$ & $0.21$ \\
    \cmidrule{2-16}
         & \textsc{UniEval} & $0.36$ & $0.28$ & $0.52$ & $0.42$ & $0.33$ & $0.25$ & $0.17$ & $0.14$ & $-$ & $-$ & $-$ & $-$  & $-$ & $-$ \\

    \cmidrule{2-16}
        & \textsc{G-Eval-3.5} & $\mathbf{0.63}$ & $\mathbf{0.55}$ & $0.59$ & $\underline{0.49}$ & $\underline{0.68}$ & $\mathbf{0.56}$ & $\underline{0.70}$ & $\underline{0.58}$ & $0.79$ & $0.67$ & $\mathbf{0.73}$ & $\mathbf{0.61}$ & $\mathbf{0.75}$ & $\mathbf{0.63}$ \\
        & \textbf{\textsc{OP-I-GPT-3.5}} & $\underline{0.60}$ & $\underline{0.51}$ & $\mathbf{0.61}$ & $\mathbf{0.51}$ & $\mathbf{0.69}$ & $\mathbf{0.56}$ & $\mathbf{0.71}$ & $\mathbf{0.59}$ & $\underline{0.80}$ & $\underline{0.68}$ & $\mathbf{0.73}$ & $\mathbf{0.61}$ & $\underline{0.74}$ & $\underline{0.61}$ \\
    \cmidrule{2-16}
        & \textsc{G-Eval-Mistral} & $0.50$ & $0.43$  & $0.54$ & $0.45$ & $0.52$ & $0.42$ & $0.54$ & $0.44$ & $0.61$ & $0.49$ & $0.55$ & $0.46$ & $0.62$ & $0.50$ \\
        & \textbf{\textsc{OP-Mistral}} &  $0.38$ & $0.32$ & $0.58$ & $0.47$ & $0.56$ & $0.45$ & $0.57$ & $0.46$ & $0.80$ & $0.67$ & $0.60$ & $0.49$ & $0.75$  & $0.62$\\
         &  \textbf{\textsc{OP-I-Mistral}} & $0.54$ & $0.45$ & $0.58$ & $0.47$ & $0.59$ & $0.47$ & $0.63^*$ & $0.51^*$ & $\mathbf{0.82}^*$  & $\mathbf{0.70}^*$ & $\mathbf{0.73}^*$ & $\mathbf{0.61}^*$ & $0.71^*$ & $0.58^*$ \\
    \bottomrule
    \end{tabular}
    }
    \caption{\textit{Spearman} ($\rho$) and \textit{Kendall Tau} ($\tau$) correlations at summary-level on $7$ dimensions for the \textsc{SummEval-Op} dataset (Round-I). For closed-source, \textsc{Op-I-Prompt} performs comparably to \textsc{G-Eval}, whereas for open-source it outperforms alternatives. $*$ represents significant performance (p-value $< 0.05$) to \textsc{G-Eval-Mistral} computed using Mann-Whitney U Test. \textsc{Humans}- averaged correlation of each annotator with the overall averaged ratings.}
    \label{tab:main_results_table}
\end{table*}

\begin{table}[t]
    \centering
    \resizebox{\columnwidth}{!}{%
    \begin{tabular}{p{\columnwidth}}
    \toprule
        \small{\textbf{\textsc{G-Eval} Response}}\\
    \midrule
    \small{<score>5</score> Explanation:
The summary covers all the major aspects discussed in the reviews, including the comfort and break-in period of the boots. It also mentions that the boots take a little while to break in, which is a common theme among the reviews. Therefore, the summary receives a high rating for its aspect coverage.}\\
    \midrule
    \small{\textbf{\textsc{Op-I-Prompt} Response}}\\
    \midrule
    \small{The metric for evaluation is Aspect Coverage. The summary should cover all the aspects that are majorly being discussed in the reviews. The reviews mention the following aspects: comfort and break-in period. The summary states "I love Frye boots. They are the most comfortable boots I have ever worn." This covers the comfort aspect. The summary also states "They do take a little while to break in, but they are worth the wait." This covers the break-in period aspect. However, the summary does not mention anything about the durability, fit and appearance aspects. Since these aspects were majorly being discussed in the reviews, the summary does not follow the metric completely. Score- <score>2</score> The summary only follows the metric to a limited extent while generating the summary from the reviews.}\\
    \bottomrule
    \end{tabular}%
    }
    \caption{\textbf{Comparative Analysis.} Sample responses generated by our \textsc{Op-I-Prompt} compared to \textsc{G-Eval} for evaluating a sample summary for the dimension \asp. On average, the scores assigned for this summary by annotators, \textsc{G-Eval}, and \textsc{Op-I-Prompt} are $2.33$, $4.09$, and $2.72$ respectively. } 
    \label{Table: example_responses}
\end{table}

\begin{table}[t]
    \centering
    \resizebox{1\columnwidth}{!}{%
    \begin{tabular}{lccccccc}
    \toprule
         \textbf{Method} & \textbf{FL} $\uparrow$ & \textbf{CO} $\uparrow$& \textbf{RE} $\uparrow$& \textbf{FA} $\uparrow$& \textbf{AC} $\uparrow$& \textbf{SC} $\uparrow$& \textbf{SP} $\uparrow$\\

    \midrule
        {\tt Human Summaries} & $4.39$ & $4.41$ & $3.78$ & $3.98$ & $3.54$ & $3.71$ & $3.66$ \\
    \midrule
        \multicolumn{8}{c}{\textit{Pre-LLMs}}\\
    \midrule
         {\tt PlanSum} & $1.86$ & $1.94$ & $1.60$ & $1.38$ & $1.52$ & $1.59$ & $1.56$ \\
         {\tt MultimodalSum} & $4.62$ & $4.09$ & $2.63$ & $2.27$ & $2.18$ & $2.76$ & $2.43$ \\
         {\tt LS-Sum-G} & $4.76$ & $4.40$ & $2.87$ & $2.74$ & $2.32$ & $3.03$ & $2.69$ \\
     \midrule
        \multicolumn{8}{c}{\textit{LLMs}}\\
    \midrule
        \chatgpt{3.5} & $\underline{4.89}$ & $4.58$ & $\mathbf{4.25}$ & $4.71$ & $4.22$ & $4.16$ & $3.96$ \\
        \gpt{4} & $\mathbf{5.00}$ & $\mathbf{4.91}$ & $3.52$ & $\mathbf{4.96}$ & $\mathbf{4.93}$ & $\mathbf{4.83}$ & $\mathbf{4.57}$ \\
        \llamatwo{2}{7} & $4.79$ & $4.34$ & $3.77$ & $4.49$ & $3.67$ & $3.79$ & $3.46$ \\
        \llamatwo{2}{13} & $4.87$ & $4.49$ & $\mathbf{4.25}$ & $4.62$ & $4.02$ & $4.00$ & $3.94$ \\
        \mistral{7} & $4.86$ & $4.60$ & $\underline{4.33}$ & $4.66$ & $\underline{4.56}$ & $4.35$ & $4.25$ \\
        \vicuna{7} & $4.83$ & $4.23$ & $3.92$ & $4.35$ & $3.96$ & $3.92$ & $3.67$ \\
        \vicuna{13} & $4.87$ & $4.41$ & $4.09$ & $4.43$ & $4.03$ & $4.00$ & $3.77$ \\
        \solar{10.7} & $\underline{4.89}$ & $\underline{4.73}$ & $4.20$ & $\underline{4.72}$ & $4.50$ & $\underline{4.56}$ & $\underline{4.35}$ \\
        \zephyr{7} & $\underline{4.89}$ & $4.36$ & $4.08$ & $4.54$ & $4.18$ & $3.95$ & $3.83$ \\
    \bottomrule
    \end{tabular}
    }
    \caption{Model-wise averaged annotator ratings of opinion summaries along $7$ dimensions (Round-II). Best scores are in \textbf{bold}, second-best are \underline{underlined}.}
    \label{tab:model_performance}
\end{table}

\section{Experiments}
We discuss the available benchmark dataset for opinion summary evaluation, the summarization models used for opinion summary generation, baseline metrics, and the implementation details.


\subsection{Summarization Models}\label{summarization_models}
\noindent\textbf{Pre-LLMs}: For the Pre-LLMs, we obtain the publicly available summaries for the Amazon test set of these models. These models were trained in a self-supervised manner using only reviews data. (\textit{1}) {\tt{\textbf{PlanSum}}} \citep{amplayo-etal-2021-unsupervised} uses content plans to create relevant review-summary pairs. The content plans take the form of aspect and sentiment distributions which are used along with input reviews for generating summaries. (\textit{2}) {\tt{\textbf{MultimodalSum}}} \citep{im-etal-2021-self} uses non-text data such as image and metadata along with reviews to generate opinion summaries. It uses a separate encoder for each modality and uses synthetic datasets to train the model in an end-to-end fashion. (\textit{3}) \citet{siledar-etal-2023-synthesize} uses lexical and semantic similarities to create a highly relevant synthetic dataset of review-summary pairs. This is then used to fine-tune any pre-trained language model for generating opinion summaries. (hereby referred to as {\tt \textbf{LS-Sum-G}}).

\noindent\textbf{LLMs}: For the LLMs, we use simple prompts\footnote{Check \textbf{Appendix \ref{appendix_sum_prompt}} for the prompt} 
to generate opinion summaries. These models were not specifically fine-tuned for opinion summarization. We use the HuggingFace library \citep{wolf2020huggingfaces} to access these models. (\textit{1}) \bchatgpt{3.5} and \bgpt{4} \citep{openai2023} are closed-source models from OpenAI optimized for dialog. We use the {\tt gpt-3.5-turbo-0125} and {\tt gpt-4-0125-preview} versions. (\textit{2}) \bllamatwo{2}{7} and \bllamatwo{2}{13} \citep{touvron2023llama} are open-source fine-tuned model from Meta with $7$B and $13$B parameters respectively. They were trained autoregressively using around $2$T tokens. We use the chat version: {\tt meta-llama/Llama-2-7b-chat-hf} model and {\tt meta-llama/Llama-2-13b-chat-hf} from the HuggingFace library. (\textit{3}) \bmistral{7} \citep{jiang2023mistral} is a $7$B instruction-tuned LLM created by MistralAI. We use the instruct version: {\tt mistralai/Mistral-7B-Instruct-v0.2} model. (\textit{4}) \bvicuna{7} and \bvicuna{13} \citep{vicuna2023} are open-source $7$B and $13$B parameter chat models trained by fine-tuning {\tt LLaMA2} on $125$K user-shared conversations collected from ShareGPT \cite{sharegpt}. We use the: {\tt lmsys/vicuna-7b-v1.5} model and {\tt lmsys/vicuna-13b-v1.5} model. (\textit{5}) \bsolar{10.7} \citep{kim2023solar} is an LLM with $10.7$B parameters, showing remarkable performance for models with parameters under $30$B. We use the version: {\tt upstage/SOLAR-10.7B-Instruct-v1.0} model. (\textit{6}) \bzephyr{7} \citep{tunstall2023zephyr} is an open-sourced fine-tuned version of {\tt mistralai/Mistral-7B-v0.1} that was trained on a mix of publicly available, synthetic datasets using Direct Preference Optimization (DPO) \citep{rafailov2023direct}. We use the beta version: {\tt HuggingFaceH4/zephyr-7b-beta} model.

\begin{figure*}[t]
    \centering
    \includegraphics[width=2\columnwidth]{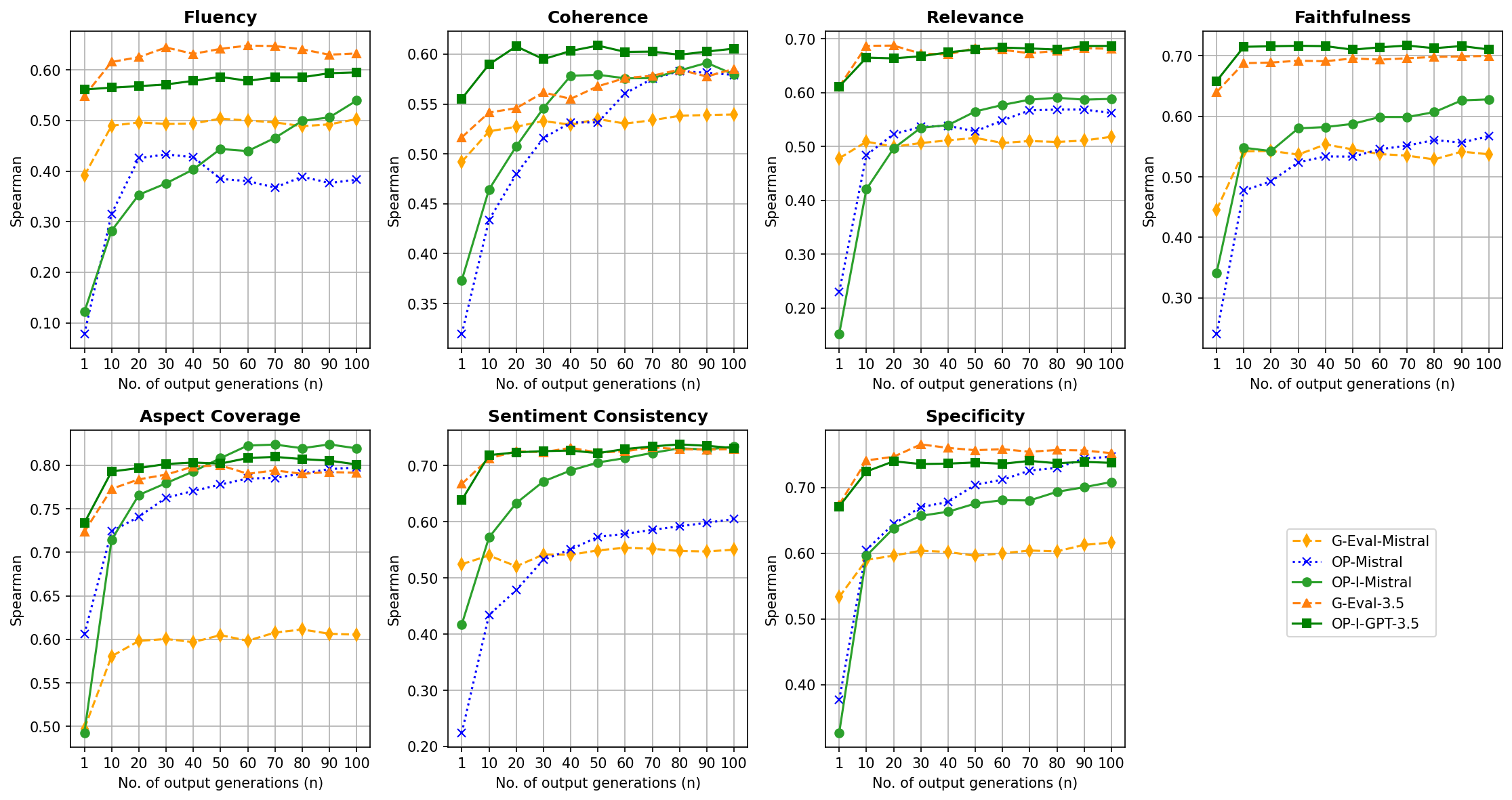}
    \caption{\textbf{\textit{Spearman} correlation scores at different number of output generations ({\tt n}) for the $7$ dimensions.} \textsc{G-Eval-3.5} and \textsc{Op-I-GPT-3.5} use the \textsc{G-Eval} and \textsc{Op-I-Prompt} respectively, with closed-source \chatgpt{3.5} as their LLM. \textsc{G-Eval-Mistral}, \textsc{Op-I-Mistral}, and \textsc{Op-Mistral} use the \textsc{G-Eval}, \textsc{Op-I-Prompt}, and \textsc{Op-Prompts}  respectively, with open-source \mistral{7} as their LLM. Generally, \textsc{Op-I-Prompt} shows better relative performance on both closed-source and open-source models.} 
    \label{fig:generation_runs}
\end{figure*}

\subsection{Baselines}
Following baseline metrics are used: \textsc{Rouge-\{1,2,L\}} score  \citep{lin-2004-rouge}, \textsc{BERTScore} \citep{Zhang2019BERTScoreET}, \textsc{BARTScore} \citep{NEURIPS2021_e4d2b6e6}, \textsc{SummaC} \citep{laban-etal-2022-summac}, \textsc{UniEval} \citep{zhong-etal-2022-towards}. We include \textsc{G-Eval} \citep{liu-etal-2023-g} as our prompt-based baseline. \textsc{G-Eval-3.5} and \textsc{G-Eval-Mistral} use \chatgpt{3.5} and \mistral{7} as their LLMs.




\subsection{Implementation Details}

For evaluation, we used \mistral{7} ({\tt mistralai/Mistral-7B-Instruct-v0.2}) as our evaluator model. We chose \mistral{7} for these reasons: (\textit{a}) it ranked best amongst the open-source models on the {\tt lmsys/chatbot-arena-leaderboard}, (\textit{b}) we found its instruction following-ness to be better than alternatives, and (\textit{c}) its {\tt $7$B} size ensures easy replication. We set the hyperparameters to \verb|n=100, temperature=0.7| to sample multiple generations. Example prompts are in \textbf{Appendix \ref{appendix_prompts}}.

\section{Results and Analysis} \label{results_analysis}

\noindent \textbf{\textsc{G-Eval} vs. \textsc{Op-I-Prompt} vs. \textsc{Op-Prompts.}} Table \ref{tab:main_results_table} and Table \ref{tab:opinsummeval_results} report the summary-level
correlation scores on the \textsc{SummEval-Op} and \textsc{OpinSummEval} dataset. In the case of closed-source models, we observe that our \textsc{Op-I-GPT-3.5} outperforms or performs comparably to \textsc{G-Eval-3.5} across all dimensions on both datasets. Specifically, our \textsc{Op-I-GPT-3.5} outperforms \textsc{G-Eval-3.5} on all $4$ dimensions for the \textsc{OpinSummEval} dataset, whereas for the \textsc{SummEval-Op} dataset, outperforms on \coh, \fa, and \asp, performs comparably on \rel \:and \sent, underperforms slightly on \fl \: and \spec.

For open-source models, overall, we observe that \textsc{Op-I-Mistral} performs the best, followed by \textsc{Op-Mistral} and then \textsc{G-Eval-Mistral}.
Figure \ref{fig:generation_runs} shows the performance of different prompt approaches over {\tt n=100} generations for $7$ dimensions. As we increase the number of generations we generally observe an improvement in the correlation. \textsc{Op-I-Mistral} shows an improvement against \textsc{G-Eval-Mistral} across all $7$ dimensions and by a large margin specifically for \asp, \sent, and \spec.

\begin{table}[t]
    \centering
    \resizebox{1\columnwidth}{!}{%
    \begin{tabular}{llccc}
         \toprule
        & & \textbf{AVG-S} &\textbf{MW} $\downarrow$ & \textbf{TT} $\downarrow$\\
        
        \midrule
        \multirow{2}{*}{\textbf{FL}} & \textbf{\textsc{G-Eval-Mistral}}$^*$ & $\mathbf{0.48}$ & \multirow{2}{*}{$\mathbf{2.9 \times 10^{-4}}$} & \multirow{2}{*}{$\mathbf{2.6 \times 10^{-4}}$} \\
         & \textsc{Op-I-Mistral} & $0.38$ &  &  \\
         
        \midrule
        \multirow{2}{*}{\textbf{CO}} & \textbf{\textsc{G-Eval-Mistral}}$^*$ & $\mathbf{0.52}$ & \multirow{2}{*}{$\mathbf{2.1 \times 10^{-4}}$} & \multirow{2}{*}{$\mathbf{3.2 \times 10^{-8}}$} \\
         & \textsc{Op-I-Mistral} & $0.47$ &  &  \\

        \midrule
        \multirow{2}{*}{\textbf{RE}} & \textsc{G-Eval-Mistral} & $\mathbf{0.51}$ & \multirow{2}{*}{$6.7 \times 10^{-2}$} &  \multirow{2}{*}{$\mathbf{4.6 \times 10^{-2}}$}\\
         & \textsc{Op-I-Mistral} & $0.49$ &  &  \\
         
        \midrule
        \multirow{2}{*}{\textbf{FA}} & \textsc{G-Eval-Mistral} & $0.53$ & \multirow{2}{*}{$\mathbf{1.9 \times 10^{-1}}$} & \multirow{2}{*}{$4.7 \times 10^{-1}$} \\
         & \textsc{Op-I-Mistral} & $\mathbf{0.54}$ &  &  \\
         
        \midrule
        \multirow{2}{*}{\textbf{AC}} & \textsc{G-Eval-Mistral} & $0.58$ & \multirow{2}{*}{$\mathbf{2.1 \times 10^{-4}}$} & \multirow{2}{*}{$\mathbf{1.9 \times 10^{-14}}$} \\
         & \textbf{\textsc{Op-I-Mistral}}$^*$ & $\mathbf{0.74}$ &  &  \\
         
        \midrule
        \multirow{2}{*}{\textbf{SC}} & \textsc{G-Eval-Mistral} & $0.54$ & \multirow{2}{*}{$\mathbf{2.1 \times 10^{-4}}$} & \multirow{2}{*}{$\mathbf{1.4 \times 10^{-7}}$} \\
         & \textbf{\textsc{Op-I-Mistral}}$^*$ & $\mathbf{0.63}$ &  &  \\
         
        \midrule
        \multirow{2}{*}{\textbf{SP}} & \textsc{G-Eval-Mistral} & $0.59$ &\multirow{2}{*}{$\mathbf{7.4 \times 10^{-4}}$}  & \multirow{2}{*}{$\mathbf{3.0 \times 10^{-4}}$} \\
         & \textbf{\textsc{Op-I-Mistral}}$^*$ & $\mathbf{0.63}$ &  &  \\
        \bottomrule
    \end{tabular}
    }
    \caption{\textbf{Significance Test.} P-values computed using Mann-Whitney U Test (MW) and T-Test (TT) between the average Spearman correlation scores (AVG-S) taken over $10$ independent generations from \textsc{G-Eval-Mistral} and \textsc{Op-I-Mistral}. \textbf{Bold} for AVG-S indicates better performance, and for MW and TT indicates p-value $< 0.05$. $*$ represents significant performance.}
    \label{tab:significance_test}
\end{table}

\noindent\textbf{Comparative Analysis.} Table \ref{Table: example_responses} shows the model responses for \textsc{G-Eval} and \textsc{Op-I-Prompt} using the \mistral{7} model. We observe the following: (\textit{a}) in general \textsc{G-Eval} erroneously assigns a higher score on average compared to the \textsc{Op-I-Prompt}, and (\textit{b}) \textsc{Op-I-Prompt} ensures that the responses adhere to a specific structure: initially outlining what is addressed and what is absent, followed by an analysis to determine the appropriate score. In contrast, the \textsc{G-Eval} responses, while mentioning various aspects covered, overlook what is missing, resulting in erroneous higher score assignments.

\noindent\textbf{Significance Testing.} We perform significance testing using the Mann-Whitney U Test \citep{mcknight2010mann} for comparison between \textsc{Op-I-Mistral} and \textsc{G-Eval-Mistral}. Table \ref{tab:main_results_table} report results for Spearman and Kendall Tau scores computed by using the scoring function with {\tt n=100}. \textsc{Op-I-Mistral} significantly (p-value $< 0.05$) outperforms \textsc{G-Eval-Mistral} on \fa, \asp, \sent, and \spec. Additionally, we group the $100$ generations into $10$ independent groups and compute Spearman correlations for each group. Table \ref{tab:significance_test} reports the Mann-Whitney U Test and T-Test p-values and arrives at a similar observation of \textsc{Op-I-Mistral} significantly outperforming on aforementioned dimensions (except \fa).

\noindent \textbf{Models for Opinion Summarization.} Table \ref{tab:model_performance} reports averaged annotator ratings for the $7$ dimensions for each model (Refer to Figure \ref{fig:model_performance_graph} for a graphical view). Overall, \gpt{4} ranks the best, followed by \solar{10.7} and \mistral{7} ranking second-best, followed by \chatgpt{3.5}. As expected, Pre-LLM models are rated the worst. These are self-supervised models and do not enjoy the liberty of being trained on trillions of tokens. All the LLMs outperform human summaries. However, because these summaries were written in the first person and as a review itself to cater to the needs when the test set was created, it is inconclusive if the LLMs outperform humans in general. We observe that \gpt{4} model scores poorly in \rel \:dimension. This we figured was due to the tendency of \gpt{4} model to try to cover every detail in the summary. Finally, \solar{10.7} and \mistral{7} with just $10.7$B and $7$B parameters respectively, outperformed \chatgpt{3.5} and comes close in performance to \gpt{4}.

\noindent \textbf{Metric Evaluation.} Reference-based metrics (\textsc{Rouge {1,2,L}}, \textsc{BERTScore}) as expected show weak correlation with human judgments. Reference-free metrics such as \textsc{BARTScore} does very poorly, however, \textsc{SummaC} performs moderately well. \textsc{UniEval} does well in \coh \:but still trails behind prompt-based approaches. \textit{To summarize, reference-based metrics are inadequate for assessing model performances in the LLMs era.}

\noindent\textbf{How sensitive is \textsc{Op-I-Prompt}?} We test \textsc{Op-I-Prompt} for $3$ definition variations of the \asp \:dimension. We paraphrase the original definition (Section \ref{metrics}) to create $2$ additional versions, ensuring the meaning is preserved (\textbf{Appendix \ref{appendix_dimension}}). We let the \textsc{Op-I-Mistral} generate {\tt n=100} responses to estimate the score using the scoring function (Section \ref{scoring_func}). The Spearman correlations for the $3$ variations are $0.82$ (Table \ref{tab:main_results_table}), $0.82$, and $0.81$, indicating that \textit{\textsc{Op-I-Prompt} is indifferent to the variations of dimensions' definition.} 

\section{Conclusion \& Future Work}
In this work, we present the \textsc{SummEval-Op} dataset, \textsc{Op-I-Prompt} and \textsc{Op-Prompts} for opinion summary evaluation on $7$ dimensions. Experimentally, we observe \textsc{Op-I-Prompt} outperforms alternatives on open-source models and performs comparably better on closed-source models showing good correlations with human judgements. Some key takeaways are: (\textit{a}) Prompts that do well for powerful closed-source LLMs may not work well for open-source LLMs; 
(\textit{b}) Opinion summaries by LLMs are preferred by humans compared to reference and previous model summaries; (\textit{c}) Reference-based summaries and metrics are inadequate in assessing LLM-based outputs. 

In the future, we plan to investigate LLMs as evaluators for performing large-scale (all reviews) and multi-source opinion summary evaluations.  

\section*{Limitations}

\begin{enumerate}
    \item We do not use \gpt{4} for evaluation purpose due to cost constraints. The primary aim of our work was to design prompts and test their applicability to both open-source and closed-source. We use \chatgpt{3.5} as the closed-source model to perform our experiments. 
    \item Our \textsc{Op-I-Prompt} was specifically designed to evaluate any dimension of the opinion summaries where \textsc{Op-Prompts} are dimension-dependent.  However, their applicability to other tasks needs further investigation and appropriate changes to the prompt. 
    \item Due to the nature of the available test sets and for benchmarking the already available models, \textsc{SummEval-Op} considers only 8 input reviews following the literature. This we believe is a major limitation in the opinion summarization field. Datasets with a larger number of reviews need to be considered for the creation of future benchmark datasets.
    \item The assessment quality of all the prompt approaches needs to be investigated for a larger amount of reviews as well.
\end{enumerate}

\section*{Acknowledgements}
We express our gratitude to the anonymous reviewers for their valuable feedback. We also extend our thanks to the annotators for their diligent and honest efforts.

\section*{Ethical Considerations}
The \textsc{SummEval-Op} dataset was created using the already available Amazon test set. We hired $3$ raters who have written papers on opinion summarization (1) or are working in the opinion summarization domain (2). These were male Masters' students aged 21-30. All the raters received stipends suitable for the tasks. 

The \textsc{Op-I-Prompt} and \textsc{Op-Prompts} are designed to offer automatic evaluation of opinion summaries for multiple dimensions. Its primary aim is to assist researchers, developers, and other stakeholders in accurately assessing summaries generated by NLG systems. However, there are potential risks associated with these prompts if they fail to accurately evaluate the quality of opinion summaries or exhibit a bias towards LLM-created content.  We urge the research community to use these prompts with caution and check their reliability for their use cases.

\bibliography{anthology,custom}

\appendix
\section{Available Benchmark Dataset}

\noindent\textbf{\textsc{OpinSummEval}}: \cite{shen2023opinsummeval} used the Yelp test set \citep{pmlr-v97-chu19b} to annotate for $4$ dimensions: {\tt readability}, {\tt self-coherence}, {\tt aspect relevance}, and {\tt sentiment consistency}. The dataset contains a total of $100$ products with $8$ reviews and $14$ different model summaries per product. Each summary was rated by $2$ annotators on $4$ dimensions. For consistency, we hereby refer to the above-mentioned dimensions as \fl, \coh, \asp, and \sent \; respectively, in line with our definitions.

\section{Rater Agreement}\label{rater_agreement}
Table \ref{tab:rmse} reports pairwise root mean squared error scores for the $3$ raters. For Round-I, we observe a difference of more than $1$ on average. For Round-II, as expected the average difference between any two ratings come down to below $1$. Table \ref{tab:rater_correlation} reports the pairwise correlations between raters as well as the correlation between each rater and average ratings for both Round-I and Round-II.

\begin{table}[t]
    \centering
    \resizebox{\columnwidth}{!}{%
    \begin{tabular}{clcccccccc}
    \toprule
          & & \multicolumn{2}{c}{\textbf{FL} $\uparrow$}  & \multicolumn{2}{c}{\textbf{CO} $\uparrow$}  & \multicolumn{2}{c}{\textbf{AC} $\uparrow$} & \multicolumn{2}{c}{\textbf{SC} $\uparrow$} \\
         \cmidrule(lr){3-4} \cmidrule(lr){5-6} \cmidrule(lr){7-8} \cmidrule(lr){9-10}
         & & $\rho$ & $\tau$ & $\rho$ & $\tau$ & $\rho$ & $\tau$ & $\rho$ & $\tau$ \\  
    \midrule
    \multirow{12}{*}{\rot{\textsc{\textbf{OpinSummEval}}}} 
        & \textsc{Rouge-1}$^\dagger$ & $0.08$ & $0.06$ & $0.11$ & $0.09$ & $0.14$ & $0.11$ & $0.00$ & $0.00$ \\
        & \textsc{Rouge-2}$^\dagger$ & $0.13$ & $0.10$ & $0.13$ & $0.11$ &  $0.15$ & $0.11$ & $0.04$ & $0.04$ \\
        & \textsc{Rouge-L}$^\dagger$ & $0.13$ & $0.10$ & $0.18$ & $0.15$ &  $0.18$ & $0.14$ & $0.07$ & $0.05$ \\
    \cmidrule{2-10}
        & \textsc{BERTScore}$^\dagger$ & $0.38$ & $0.30$ & $0.20$ & $0.17$ & $0.20$ & $0.16$ & $0.08$ & $0.06$ \\
        & \textsc{BARTScore}$^\dagger$ & $0.42$ & $0.33$ & $0.35$ & $0.29$ & $0.28$ & $0.22$ & $0.41$ & $0.34$ \\
        & \textsc{SummaC}$^\dagger$ & $0.06$ & $0.05$ & $0.02$ & $0.01$ & $0.07$ & $0.06$ & $0.20$ & $0.17$ \\
    \cmidrule{2-10}
    & \textsc{ChatGPT-3.5}$^\dagger$ & $0.47$ & $0.42$ & $0.28$ & $0.25$ & $\underline{0.34}$ & $\underline{0.30}$ & $0.37$ & $0.33$\\  
    & \textsc{G-Eval-3.5}$^\dagger$ & $0.41$ & $0.36$ & $0.29$ & $0.26$ & $0.27$ & $0.23$ & $0.38$ & $0.34$\\
    & \textbf{\textsc{Op-I-GPT-3.5}} & $\mathbf{0.51}$ & $\mathbf{0.46}$ & $0.36$ & $0.32$ & $0.33$ & $0.29$ & $0.43$ & $0.39$\\
    \cmidrule{2-10}
    & {\textsc{G-Eval-Mistral}} & $\underline{0.46}$ & $\underline{0.41}$ & $\underline{0.41}$ & $\underline{0.38}$ & $\underline{0.36}$ & $\underline{0.32}$ & $\mathbf{0.49}$ & $\mathbf{0.45}$ \\ 
    & \textbf{\textsc{OP-Mistral}} & $0.35$ & $0.33$ & $\mathbf{0.45}$ & $\mathbf{0.41}$ & $0.34$ & $0.30$ & $0.45$ & $0.41$\\   
    &  \textbf{\textsc{OP-I-Mistral}} & $\underline{0.46}$ & $\underline{0.41}$ & $0.37$ & $0.35$ & $\mathbf{0.38}$  & $\mathbf{0.34}$ & $\mathbf{0.49}$ & $\mathbf{0.45}$ \\
    \bottomrule
    \end{tabular}
    }
    \caption{\textit{Spearman} ($\rho$) and \textit{Kendall Tau} ($\tau$) correlations at summary-level on $7$ dimensions for the \textsc{OpinSummEval} dataset. For closed-source, \textsc{Op-I-Prompt} performs comparably to \textsc{G-Eval}, whereas for open-source it outperforms alternatives. $\dagger$ represents results as reported in \citet{shen2023opinsummeval}}
    \label{tab:opinsummeval_results}
\end{table}

\begin{table}[t]
    \centering
    \resizebox{1\columnwidth}{!}{%
    \begin{tabular}{lcccccccc}
         \toprule
        & \textbf{FL} $\downarrow$ & \textbf{CO} $\downarrow$ & \textbf{RE} $\downarrow$ & \textbf{FA} $\downarrow$ & \textbf{AC} $\downarrow$ & \textbf{SC} $\downarrow$ & \textbf{SP} $\downarrow$\\
        \midrule
        \multicolumn{8}{c}{\textit{Round-I}}\\
        \midrule
        \textbf{A1-A2} & $0.95$ & $1.06$ & $1.01$ & $1.09$ & $0.91$ & $1.08$ & $0.95$ \\
        \textbf{A2-A3} & $0.44$ & $0.86$ & $1.09$ & $1.05$ & $0.84$ & $1.19$ & $1.42$ \\
        \textbf{A1-A3} & $1.00$ & $1.23$ & $1.16$ & $1.24$ & $1.15$ & $1.47$ & $1.55$ \\
        \midrule
        \textbf{AVG-I} & $0.80$ & $1.05$ & $1.09$ & $1.13$ & $0.97$ & $1.25$ & $1.31$ \\
        \midrule
        \multicolumn{8}{c}{\textit{Round-II}}\\
        \midrule
        \textbf{A1-A2} & $0.55$ & $0.66$ & $0.65$ & $0.60$ & $0.64$ & $0.64$ & $0.60$ \\
        \textbf{A2-A3} & $0.31$ & $0.62$ & $0.67$ & $0.67$ & $0.68$ & $0.71$ & $0.79$ \\
        \textbf{A1-A3} & $0.53$ & $0.73$ & $0.67$ & $0.68$ & $0.76$ & $0.76$ & $0.73$ \\
        \midrule
        \textbf{AVG-II} & $0.47$ & $0.67$ & $0.66$ & $0.65$ & $0.69$ & $0.70$ & $0.71$ \\
        \bottomrule
    \end{tabular}
    }
    \caption{\textbf{Round-I and Round-II Ratings:} Pairwise \textit{Root Mean Squared Error} scores for $3$ raters A1, A2, and A3.}
    \label{tab:rmse}
\end{table}

\begin{table*}[t]
    \centering
    \resizebox{2\columnwidth}{!}{%
    \begin{tabular}{clcccccccccccccc}
    \toprule
          & & \multicolumn{2}{c}{\textbf{FL} $\uparrow$}  & \multicolumn{2}{c}{\textbf{CO} $\uparrow$} & \multicolumn{2}{c}{\textbf{RE} $\uparrow$} & \multicolumn{2}{c}{\textbf{FA} $\uparrow$} & \multicolumn{2}{c}{\textbf{AC} $\uparrow$} & \multicolumn{2}{c}{\textbf{SC} $\uparrow$} & \multicolumn{2}{c}{\textbf{SP} $\uparrow$}\\
         \cmidrule(lr){3-4} \cmidrule(lr){5-6} \cmidrule(lr){7-8} \cmidrule(lr){9-10} \cmidrule(lr){11-12} \cmidrule(lr){13-14} \cmidrule(lr){15-16}
         & & $\rho$ & $\tau$ & $\rho$ & $\tau$ & $\rho$ & $\tau$ & $\rho$ & $\tau$ & $\rho$ & $\tau$ & $\rho$ & $\tau$ & $\rho$ & $\tau$ \\  
   
    \midrule
        \multicolumn{16}{c}{\textit{Pairwise correlation among raters}}\\
    \cmidrule{2-16}
        \multirow{10}{*}
        {\textit{\rot{Round-I}}} & \textbf{A1-A2} & $0.58$ & $0.56$ & $0.54$ & $0.50$ & $0.65$ & $0.60$ & $0.73$ & $0.68$ & $0.78$ & $0.72$ & $0.60$ & $0.53$ & $0.65$ & $0.59$ \\
        & \textbf{A2-A3} & $0.79$ & $0.78$ & $0.40$ & $0.38$ & $0.52$ & $0.47$ & $0.63$ & $0.58$ & $0.77$ & $0.71$ & $0.56$ & $0.51$ & $0.58$ & $0.53$ \\
        & \textbf{A1-A3} & $0.55$ & $0.53$ & $0.34$ & $0.31$ & $0.40$ & $0.36$ & $0.60$ & $0.54$ & $0.74$ & $0.68$ & $0.57$ & $0.51$ & $0.57$ & $0.51$ \\
    \cmidrule{2-16}
        & \textbf{AVG-I} & $0.64$ & $0.62$ & $0.43$ & $0.40$ & $0.52$ & $0.48$ & $0.65$ &$ 0.60$ & $0.76$ & $0.70$ & $0.58$ & $0.52$ & $0.60$ & $0.54$\\
     \cmidrule{2-16}
    
        \multicolumn{16}{c}{\textit{Pairwise correlation between raters and the overall average ratings}}\\
    \cmidrule{2-16}
        & \textbf{A-A1} & $0.95$ & $0.93$ & $0.86$ & $0.82$ & $0.81$ & $0.74$ & $0.86$ & $0.80$ & $0.91$ & $0.85$ & $0.85$ & $0.77$ & $0.87$ & $0.80$ \\
        & \textbf{A-A2} & $0.70$ & $0.67$ & $0.72$ & $0.67$ & $0.82$ & $0.75$ & $0.81$ & $0.75$ & $0.91$ & $0.85$ & $0.85$ & $0.78$ & $0.87$ & $0.80$ \\
        & \textbf{A-A3} & $0.57$ & $0.54$ & $0.56$ & $0.51$ & $0.74$ & $0.67$ & $0.81$ & $0.76$ & $0.88$ & $0.81$ & $0.76$ & $0.69$ & $0.76$ & $0.69$ \\
        \cmidrule{2-16}
        & \textbf{AVG-II} & $0.74$ & $0.71$ & $0.71$ & $0.67$ & $0.79$ & $0.72$ & $0.83$ & $0.77$ & $0.90$ & $0.84$ & $0.82$ & $0.75$ & $0.83$ & $0.76$\\
        
    \midrule
        \multicolumn{16}{c}{\textit{Pairwise correlation among raters}}\\
    \cmidrule{2-16}
        \multirow{10}{*}
        {\textit{\rot{Round-II}}} & \textbf{A1-A2} & $0.63$ & $0.61$ & $0.64$ & $0.61$ & $0.80$ & $0.75$ & $0.83$ & $0.79$ & $0.85$ & $0.80$ & $0.77$ & $0.72$ & $0.81$ & $0.76$ \\
        & \textbf{A2-A3} & $0.85$ & $0.84$ & $0.59$ & $0.56$ & $0.78$ & $0.73$ & $0.77$ & $0.73$ & $0.83$ & $0.78$ & $0.77$ & $0.73$ & $0.81$ & $0.77$ \\
        & \textbf{A1-A3} & $0.66$ & $0.65$ & $0.59$ & $0.56$ & $0.77$ & $0.72$ & $0.78$ & $0.73$ & $0.84$ & $0.79$ & $0.79$ & $0.74$ & $0.82$ & $0.78$ \\
    \cmidrule{2-16}
        & \textbf{AVG-I} & $0.71$ & $0.70$ & $0.61$ & $0.58$ & $0.78$ & $0.73$ & $0.79$ & $0.75$ & $0.84$ & $0.79$ & $0.78$ & $0.73$ & $0.81$ & $0.77$ \\
    \cmidrule{2-16}
    \multicolumn{16}{c}{\textit{Pairwise correlation between raters and the overall average ratings}}\\
    \cmidrule{2-16}
        & \textbf{A-A1} & $0.94$ & $0.92$ & $0.87$ & $0.83$ & $0.91$ & $0.85$ & $0.89$ & $0.84$ & $0.94$ & $0.88$ & $0.92$ & $0.86$ & $0.92$ & $0.87$ \\
        & \textbf{A-A2} & $0.76$ & $0.74$ & $0.80$ & $0.75$ & $0.91$ & $0.86$ & $0.87$ & $0.82$ & $0.93$ & $0.88$ & $0.91$ & $0.85$ & $0.92$ & $0.87$ \\
        & \textbf{A-A2} & $0.69$ & $0.67$ & $0.76$ & $0.71$ & $0.91$ & $0.85$ & $0.92$ & $0.88$ & $0.93$ & $0.86$ & $0.91$ & $0.85$ & $0.92$ & $0.87$ \\
        \cmidrule{2-16}
        & \textbf{AVG-II} & $0.80$ & $0.77$ & $0.81$ & $0.76$ & $0.91$ & $0.86$ & $0.89$ & $0.85$ & $0.93$ & $0.87$ & $0.91$ & $0.85$ & $0.92$ & $0.87$ \\
    \bottomrule
    \end{tabular}
    }
    \caption{\textbf{Rater Correlations:} Pairwise \textit{Spearman} ($\rho$) and \textit{Kendall Tau} ($\tau$) correlations at summary-level for $3$ raters A1, A2, and A3 along with the average of their ratings A.}
    \label{tab:rater_correlation}
\end{table*}

\begin{figure*}[t]
    \centering
    \includegraphics[width=2\columnwidth]{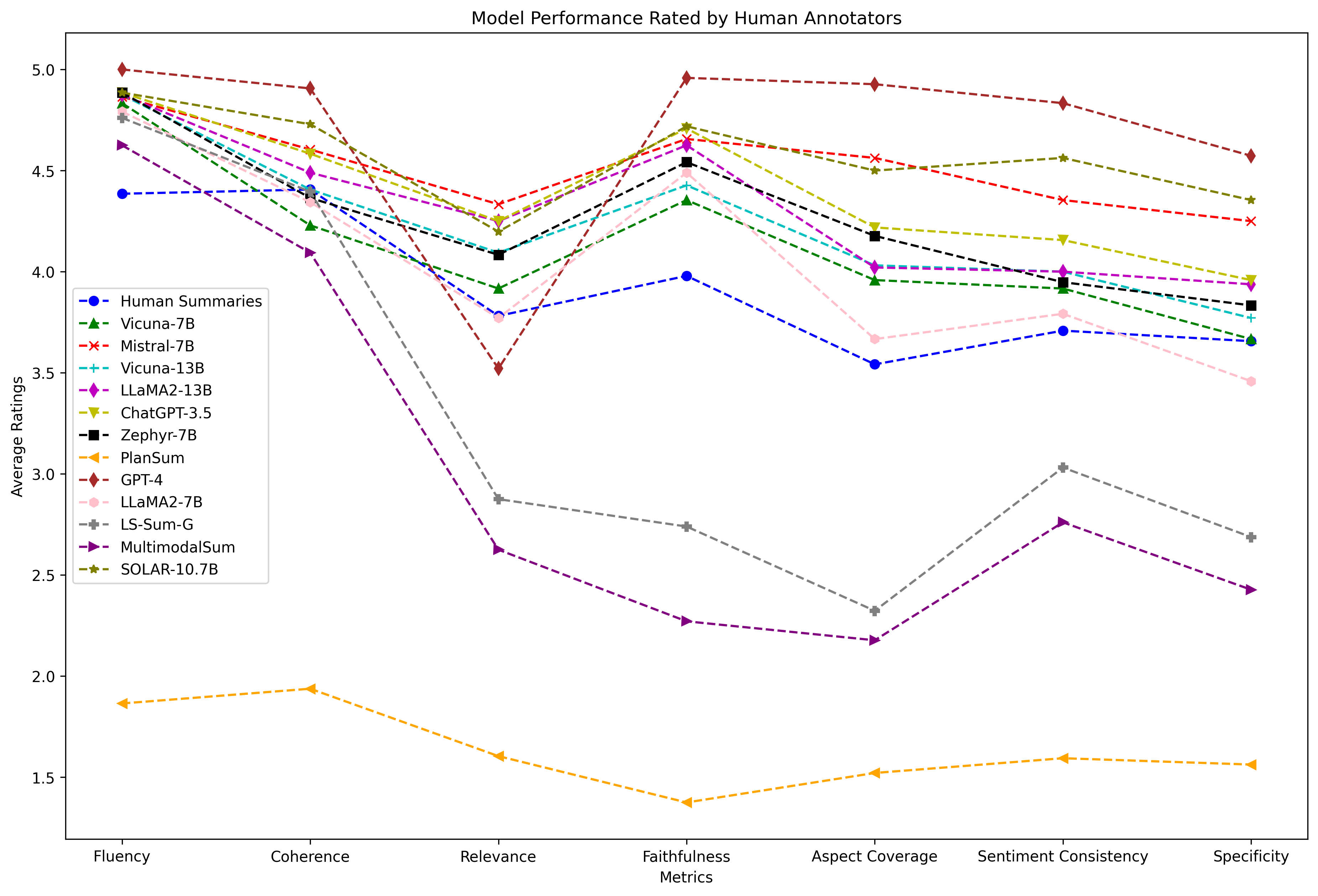}
    \caption{\textbf{Performance of different models as rated by human annotators (Round-II).} We observe that \gpt{4} performs the best followed by \solar{10.7} and \mistral{7}. Self-supervised models perform worse. In general, all the LLMs perform better than human annotated summaries.} 
    \label{fig:model_performance_graph}
\end{figure*}

\section{Prompts}\label{appendix_prompts}
For brevity, we provide different prompts for only a single dimension- Aspect Coverage. We will release prompts for all the dimensions across different approaches publicly.

\subsection{\textsc{Op-I-Prompt} for Aspect Coverage}

{\tt \textbf{Task Description}:\\
You will be given a set of reviews using which a summary has been generated. Your task is to evaluate the summary based on the given metric. Evaluate to which extent does the summary follows the given metric considering the reviews as the input. Use the following evaluation criteria to judge the extent to which the metric is followed. Make sure you understand the task and the following evaluation metric very clearly.
\\\\
\textbf{Evaluation Criteria}:\\
The task is to judge the extent to which the metric is followed by the summary.
Following are the scores and the evaluation criteria according to which scores must be assigned.\\
<score>1</score> - The metric is not followed at all while generating the summary from the reviews.\\
<score>2</score> - The metric is followed only to a limited extent while generating the summary from the reviews.\\
<score>3</score> - The metric is followed to a good extent while generating the summary from the reviews.\\
<score>4</score> - The metric is followed mostly while generating the summary from the reviews.\\
<score>5</score> - The metric is followed completely while generating the summary from the reviews.
\\\\
\textbf{Metric}:\\
Aspect Coverage - The summary should cover all the aspects that are majorly being discussed in the reviews. Summaries should be penalized if they miss out on an aspect that was majorly being discussed in the reviews and awarded if it covers all.
\\\\
\textbf{Reviews}:\\ 
\{\}
\\\\
\textbf{Summary}:\\ 
\{\}
\\\\
\textbf{Evaluation Steps}:
\\
Follow the following steps strictly while giving the response:\\
1.First write down the steps that are needed to evaluate the summary as per the metric. Reiterate what metric you will be using to evaluate the summary.\\
2.Give a step-by-step explanation if the summary adheres to the metric considering the reviews as the input. Stick to the metric only for evaluation.  \\
3.Next, evaluate the extent to which the metric is followed.\\
4.Use the previous information to rate the summary using the evaluation criteria and assign a score within the <score></score> tags.
\\\\
Note: Strictly give the score within <score></score> tags only e.g Score- <score>5</score>.
\\\\
First give a detailed explanation and then finally give a single score following the format: Score- <score>5</score>
\\\\
THE EVALUATION AND SCORE MUST BE ASSIGNED STRICTLY ACCORDING TO THE METRIC ONLY AND NOTHING ELSE!
\\\\
\textbf{Response}: }

\subsection{\textsc{Op-Prompts} for Aspect Coverage}

{\tt \textbf{Task Description}:\\
You will be given a set of reviews. You will then be given one summary written for the set of reviews. Your task is to rate the summary on one metric. Make sure you understand the following evaluation metric very clearly. Your task is to rate the summary corresponding to the given reviews on the evaluation criteria.
\\\\
\textbf{Evaluation Criteria}:\\
Aspect Coverage - The summary should cover all the aspects that are majorly being discussed in the reviews. Summaries should be penalized if they miss out on an aspect that was majorly being discussed in the reviews and awarded if it covers all.\\
<score>1</score> - Summary does not cover any important aspects present in the reviews.\\
<score>2</score> - Summary does not cover most of the important aspects present in the reviews.\\
<score>3</score> - Summary covers around half of the important aspects present in the reviews.\\
<score>4</score> - Summary covers most of the important aspects present in reviews.\\
<score>5</score> - Summary covers all the important aspects discussed in reviews.
\\\\
\textbf{Metric}:\\
Aspect Coverage - The summary should cover all the aspects that are majorly being discussed in the reviews. Summaries should be penalized if they miss out on an aspect that was majorly being discussed in the reviews and awarded if it covers all.
\\\\
\textbf{Reviews}:\\ 
\{\}
\\\\
\textbf{Summary}:\\ 
\{\}
\\\\
\textbf{Evaluation Steps}:
\\
Let's go step-by-step. Follow the following steps strictly while giving the response:\\
1.Identify the important aspects present in the reviews and list them with numbering\\
2.Identify the important aspects present in the summary and list them with numbering\\
3.Identify the important aspects covered by the summary that are present in the reviews and list them with numbering\\
4.Calculate the total number of important aspects covered by the summary that are present in the reviews \\
5.Calculate the total number of important aspects present in the reviews\\
6.Finally use the evaluation criteria to output only a single score within <score></score> tags.\\
\\
Note: Strictly give the score within <score></score> tags only e.g Score- <score>5</score>.
\\\\
First give a detailed explanation of how much is the coverage and then finally give a single score following the format: Score- <score>5</score>
\\\\
\textbf{Response}:  }

\subsection{\textsc{G-Eval} for Aspect Coverage}
{\tt \textbf{Task Description}:\\
You will be given a set of reviews and a corresponding summary. Make sure you understand the following evaluation metric very clearly. Your task is to rate the summary corresponding to the given reviews on the evaluation criteria.
\\\\
\textbf{Evaluation Criteria}:
Aspect Coverage (1-5) - The summary should cover all the aspects that are majorly being discussed in the reviews. Summaries should be penalized if they miss out on an aspect that was majorly being discussed in the reviews and awarded if it covers all.
\\\\
\textbf{Reviews}:\\
\{\}
\\\\
\textbf{Summary}:\\
\{\}
\\\\
\textbf{Evaluation Steps}:\\
1.Read through the given set of reviews carefully.\\
2.Compare the content of the reviews to the provided summary.\\
3.Evaluate whether the summary covers all the major aspects that are being discussed in the reviews.\\
4.Rate the summary on a scale of 1-5 based on how well it covers the aspects discussed in the reviews.\\
5.Provide a brief explanation for your rating, citing specific examples from the reviews and summary.
\\\\
Note: Strictly give the score within <score></score> tags only e.g Score: <score>5</score>.
\\\\
\textbf{Response}: }

\subsection{Summarization Prompt}\label{appendix_sum_prompt}
{\tt Generate a summary for the following set of reviews. Generate the summary in a paragraph format. No bulletpoints or explanations needed. Just output the summary text. 
\\\\
\textbf{Reviews}: 
\\
\{\}\\\\
\textbf{Summary}:}

\section{Dimension Definitions}\label{appendix_dimension}
For ablation, we try out three different definition variations of \asp.

\noindent\textbf{Definition 1:} The summary should cover all the aspects that are majorly being discussed in the reviews. Summaries should be penalized if they miss out on an aspect that was majorly being discussed in the reviews and awarded if it covers all.

\noindent\textbf{Definition 2:} This refers to the comprehensiveness of a summary in capturing all significant aspects discussed in reviews. A summary is evaluated based on its ability to include major topics of discussion; it is deemed deficient if it overlooks any crucial aspect and commendable if it encompasses them all.

\noindent\textbf{Definition 3:} Aspect coverage pertains to the extent to which a summary encapsulates the key facets discussed in reviews. Summaries are evaluated based on their ability to incorporate major discussion points. They are considered deficient if they omit any critical aspect and commendable if they address them all comprehensively.

\end{document}